\documentclass[sigconf]{acmart}

\AtBeginDocument{%
  \providecommand\BibTeX{{%
    \normalfont B\kern-0.5em{\scshape i\kern-0.25em b}\kern-0.8em\TeX}}}

\setcopyright{none}
\copyrightyear{2024}
\acmYear{2024}
\acmDOI{10.48550/arXiv.2406.14273}

\acmConference[CHIWORK '24]{Symposium on Human-Computer Interaction for Work}{June 25--27, 2024}{Newcastle-upon-Tyne, UK}
\acmBooktitle{Symposium on Human-Computer Interaction for Work (CHIWORK '24), June 23--27, 2024, Newcastle-upon-Tyne, UK} 

\acmISBN{}

\begin{document}

\title{The Impact of AI on Perceived Job Decency and Meaningfulness: A Case Study}

\author{Kuntal Ghosh}
\authornote{Both authors contributed equally to this research.}
\email{kuntal.ghosh@uni-siegen.de}
\affiliation{%
  \institution{University of Siegen}
  \streetaddress{Kohlbettstraße 15}
  \city{Siegen}
  \state{North-Rhine Westphalia}
  \country{Germany}
  \postcode{57072}
}

\author{Shadan Sadeghian}
\email{shadan.sadeghian@uni-siegen.de}
\affiliation{%
  \institution{University of Siegen}
  \streetaddress{Kohlbettstraße 15}
  \city{Siegen}
  \state{North-Rhine Westphalia}
  \country{Germany}
  \postcode{57072}
}

\renewcommand{\shortauthors}{Ghosh and Sadeghian}

\begin{abstract}
    The proliferation of Artificial Intelligence (AI) in workplaces stands to change the way humans work, with job satisfaction intrinsically linked to work life.
    Existing research on human-AI collaboration tends to prioritize performance over the experiential aspects of work.
    In contrast, this paper explores the impact of AI on job decency and meaningfulness in workplaces. Through interviews in the Information Technology (IT) domain, we not only examined the current work environment, but also explored the perceived evolution of the workplace ecosystem with the introduction of an AI.
    Findings from the preliminary exploratory study reveal that respondents tend to visualize a workplace where humans continue to play a dominant role, even with the introduction of advanced AIs. In this prospective scenario, AI is seen as serving as a complement rather than replacing the human workforce. Furthermore, respondents believe that the introduction of AI will maintain or potentially increase overall job satisfaction.
\end{abstract}


\received{12 March 2024}
\received[accepted]{13 April 2024}

\maketitle


\section{Introduction}
With more than one-third of the human lifespan dedicated to work, it is evident that work plays an integral role in shaping our lives, with work-life quality determining job satisfaction \cite{ahmad2013paradigms}. Technology has long played a pivotal role in the evolution of work, evolving from primitive tools to contemporary computing devices. With the introduction of AI in workplaces\cite{marquis2024proliferation}, transitions in work practices are inevitable. However, research involving human-AI collaboration tends to prioritize performance-related factors such as efficiency and effectiveness over the experiential aspects \cite{sadeghian2022artificial}. While this may be beneficial in the short term, in the long term, it might lead to workers' disengagement and dissatisfaction. Previous research shows that job satisfaction stems from two sources: job decency, and job meaningfulness \cite{blustein2023understanding}. 
Decent work outlines the basic human needs at the workplace, including fair remuneration, job stability, social dialogue, and equal opportunities and conduct \cite{office2014global, pereira2019empirical}.
Meaningful work is rather subjective, with various documented perspectives, such as holding positive meaning \cite{rosso2010meaning}, improving social welfare \cite{steger2012measuring}, answering one's calling \cite{dik2009calling, lysova2019meaningful}, and understanding one's purpose \cite{pratt2003fostering}. It includes aspects like work-impact visibility \cite{grant2007impact, grant2007relational}, workplace appreciations \cite{wrzesniewski2003interpersonal}, positive connections \cite{colbert2016flourishing}, social responsibility \cite{aguinis2019corporate}, professional growth \cite{fletcher2019can}, challenging work climate \cite{kim2020thriving}, autonomy, and skill variety \cite{hackman1976motivation}. \par
However, it is still unclear how the introduction of AI changes the perception of decency and meaningfulness, and consequently affects job satisfaction. To this end, this paper attempts to answer three research questions: \textbf{RQ1:} What factors influence the perception of a job as decent? \textbf{RQ2:} What factors contribute to job meaningfulness? \textbf{RQ3:} How does the introduction of AI influence the perception of job decency and meaningfulness? To answer these questions, we conducted semi-structured interviews with eight IT professionals and asked about their current and future perception of job decency and meaningfulness with AI. Our results show that with the introduction of AI, the overall job satisfaction is higher, with job decency improving considerably and job meaningfulness remaining the same.

\section{Method}
We conducted semi-structured interviews \cite{lazar2017research} using anticipatory ethnography method \cite{lindley2014anticipatory} with IT professionals to understand their perception of job decency and meaningfulness in the present and future of work with AI. The interviews consisted of two segments: one focusing on their present experiences and another envisioning future scenarios. Each segment had three parts: introductory, decent work, and meaningful work. In the present introductory part, we asked the participants to describe a typical working day and their work environment, technology used, and (dis)satisfying job aspects. For future scenarios, we asked them to imagine an AI and describe what working with it would be like. In the decent work part, in line with \cite{pereira2019empirical, office2014global}, we asked questions regarding their working hours, work-life balance, job security, growth opportunities, compensation, and working conditions. 
In the present segment, participants answered questions based on their current experiences, while in the future segment, they responded based on their speculation of the AI in their workplace. Likewise, in the meaningful work part, in line with \cite{aguinis2019corporate, fletcher2019can, colbert2016flourishing, wrzesniewski2003interpersonal, hackman1976motivation, kim2020thriving, grant2007impact, grant2007relational} we asked participants about their job fulfilment, social interactions, goals alignment, co-workers’ treatment, autonomy, skill variety, task stimulation, social image and workplace recognition. 
Eight male IT professionals (age: 21–40, m=31, SD=1.85) participated in this study. This group comprised software engineer(3), machine language engineer(2), deep learning engineer(1), front-end developer(1), data analyst(1). All participants had prior exposure to large language models like ChatGPT or GitHub Copilot. They were recruited through social media and word-of-mouth and were compensated with €12. The interviews were conducted online via Webex\footnote{https://www.webex.com/}. 

\section{Findings}
We transcribed the recordings using Adobe Premiere Pro 
and followed the emergent coding method based upon the grounded theory concept \cite{lazar2017research} to code our data using MAXQDA\footnote{https://www.maxqda.com/}. All eight participants envisioned their AIs in the form of assistants providing them with a \textit{"helping hand"}. AIs were depicted as physical entities, digital components, or avatars in the virtual space, with various interaction modalities, including verbal, textual, visual, neural, and haptic. 

\subsection{Decent Work (Present versus Future)}
In present scenarios, all eight participants reported having 39-40 weekly \textbf{working hours}, although the \textit{actual time worked} was slightly less. Six found it adequate, citing flexible timing and dedicated idle time, while two found it prolonged and excessive. However, all eight participants were satisfied with their \textbf{work-life balance} due to hybrid work-mode, and flexible working hours: \textit{"when I have flexibility in my work hours, then I have flexibility in my personal life"} (P8). In the future scenarios, five felt their hours would reduce with AI sharing the workload. One believed that hours would reduce only if he is \textit{"supposed to do the same amount of work after introducing the AI"} (P8), while another deemed that he would still be expected to spend the same hours on newly added tasks. Four believed that the time cutback would improve work-life balance, while three could not foresee any differences due to changing expectations at the workplace, or seeing themselves as the main executors of work.

In present scenarios, two participants said they do not have \textbf{job security} due to the economic crisis. Six felt secure due to their vital roles, favourable employment laws, organizational support, or meeting expectations. All eight mentioned having abundant \textbf{growth opportunities}, but two of them said the topics hinge towards market trends. With numerous opportunities, participants felt responsible for utilizing them: \textit{"it's on me to motivate myself to take this budget and do something meaningful instead of wasting it"} (P4). 
In the future scenarios, five felt sustained learning and development are paramount for employment stability. All eight believed that the need to learn new skills would be perpetual due to evolving demands, and there would be no dearth of growth chances. Two stated that the scope of growth would expand since AI would do repetitive tasks and they would get \textit{"more challenging [problems] which are not solved yet"}. Fear of layoff was non-existent amongst four who \textit{"expect[ed] to have the brain"} in human-AI teaming or were adamant that \textit{"AI is not here to take our jobs"}. However, three felt job security would decline as AIs become more competent with time.


All eight participants in present scenarios spoke positively about their \textbf{working conditions}, citing aspects like open-office space, flat hierarchies, post-work meetups, and supplementary amenities. \textbf{Compensation} was benchmarked against geography-specific market salaries for similar jobs, with five satisfied, and three dissatisfied. In the future scenarios, all eight felt the work environment and culture for humans would not change, but two of them expected the work facilities with AI to become highly sophisticated, and costly to maintain. Three imagined an increase in salaries due to their enhanced competencies, while one could not envision any change. Two debated on reducing salary for fewer work hours, and another believed that organizations would offset AI maintenance costs by decreasing the payroll.

\subsection{Meaningful Work (Present versus Future)}
In present scenarios, all eight participants found their \textbf{jobs fulfilling}. Having significant impact (n=5) \textit{"I feel like what we are doing is different, definitely affecting our users and making their lives better."} (P1), abundant learning opportunities (n=2), and remuneration (n=1) were named as sources for meaning.
In the future scenarios, six believed their work would still be worthwhile since humans would use the AI as \textit{"a tool"} to be more efficient. While one participant felt his job might no longer be meaningful since it could be taken over by AI, another asserted that he \textit{"would only engage in something meaningful [impacts people positively]"} (P5).

\textbf{Interaction with colleagues} in present scenarios was appreciated by seven participants, with in-person engagement preferred to remote. Acquiring different perspectives, mentoring, empathizing with work issues, or socializing are some of the aspects mentioned by participants. All eight participants praised the \textbf{treatment from coworkers}. They found their colleagues to respect boundaries and be fair, transparent, nurturing, and treat them as equals.
In the future scenarios, the collective accord was that interpersonal relationships would be more profound as conversations shift away from work-related inquiries to personal and informal topics: \textit{"maybe you will not interact about your work; maybe you can interact about other things! You can ask how was your day and all that"} (P1). Four said that due to humans' inherent social nature or since workplace ownership still rests with humans, they would prefer interacting with humans to AIs. Seven mused that human behaviour is independent of technological advancements. Four believed that workplace treatment would remain unchanged since AI would be used by all and exist only to assist them. However, one mentioned that his colleagues might perceive the AI as the principal executor of tasks and alter their behaviour towards him.



\textbf{Goals correlation} in present scenarios was appreciated by five participants whose aspirations were acknowledged by management. One reflected on the organizational tendency to coordinate employee goals with market trends.
In the future scenarios, four felt their goals would align with the organization since it involves upskilling, being more efficient, and creating impact. One argued that AI would narrow career choices. Two stated this aspect is unrelated to AI.

In present scenarios, seven participants were content with their \textbf{workplace autonomy}, even if they had to perform some predetermined tasks. However, one voiced a sense of helplessness: \textit{"you feel that your knowledge is not being used to the fullest, because if you know that there is a better solution, but you are told that - no, you have to stick to this, then you [do not feel] motivated"} (P6).
In the future scenarios, seven anticipated retaining their autonomy, if not augmenting it, as they would maintain control and tailor the AI to meet their needs: \textit{"AI will probably never be critical decision makers because humans will still be the decision maker ... the law cannot get hold of AI, someone has to be accountable"} (P5).

\textbf{Skill variety} in present scenarios required working on multiple technologies, solving novel problems, and handling complex systems, which motivated all eight participants in their daily work: \textit{"it's something which excites me"} (P6). Although everyone were satisfied with their \textbf{task stimulation}, some days would have their fair share of monotony: \textit{"it's not always stimulating, sometimes you have to work on boring stuff, but it's sufficient"} (P4).
In the future scenarios, all eight participants believed that the evolution of work would elevate organizational expectations, with new problems arising: \textit{"when the tool advances, companies will come up with more challenging [tasks]"} (P4). According to them, \textit{"challenges would always be there"}, but now they would have a \textit{"helping hand"}. Six emphasized that the novelty of unfamiliar problems would make solving them intellectually stimulating since AI would tackle all monotonous tasks. However, two stated that their mental engagement would be reduced with AIs sharing their workload.

In present scenarios, seven participants reported a positive \textbf{social image} due to their organizations' reputation, acquired soft skills, and \textit{"solving difficult tasks"}. 
All eight revealed they receive \textbf{credit and recognition} for their work, although not everything warrants commendation: \textit{"there isn't acclamation for everything you do, it doesn't happen in professional life"} (P8).
In the future scenarios, five envisaged no change in public perception since AIs would be ubiquitous and only exist to assist humans. While one believed his social status would decline due to the \textit{"misconception of the public"} that AI does his work, another considered his image on social media: \textit{"if [my] AI posts something offensive, that will definitely impact my image"} (P5). Six believed they would be appreciated due to their enhanced competencies, normalization of AI-usage \textit{"we won't look down on someone using AI because we are all using it"} (P8), workplace culture, or lack of liability within AI systems \textit{"AIs won't be credited, the accountability and ownership still stays with us"} (P5). 


\vspace{-3mm}
\section{Discussion}
Our results show that the interviewees from the IT-domain did not anticipate a decrease in their overall job satisfaction with the introduction of AI; in fact, some expected it to increase. They mentioned that AI will make their work more fulfilling, provide them with assistance in their tasks, and improve social interactions with colleagues.
In terms of decent work aspects, they imagined reduced working hours, better work-life balance, plentiful growth opportunities, and enhanced working conditions in their future workplace with AI. Although they anticipated fair compensation, they expressed concerns about potential job insecurity resulting from the evolving nature of work, including job displacement and shifts in employment patterns. 
Regarding the elements of meaningful work, they anticipated solving novel problems by utilizing AI to their advantage and achieve higher fulfilment through increased impact, enhanced competencies and elevated workplace camaraderie. However, they also recognized the possibility of being misperceived as passive observers while AI performs all the work, potentially undermining their contributions and subsequent appreciations.

Since all eight participants were familiar with AI, their outlooks might have markedly differed from individuals in other job sectors.
To address this limitation and ensure a comprehensive understanding of varied perspectives, we have planned the recruitment of participants from two other distinct worker groups: service workers, and care givers. This strategic approach will enable us to systematically explore the perceptions of job decency and meaningfulness in collaboration with AI across various professional contexts.

\bibliographystyle{ACM-Reference-Format}
\bibliography{sample-base}



\end{document}